# 3D Reconstruction of Spherical Images based on Incremental Structure from Motion

**San Jiang** [1, 2], **Kan You** [1], **Yaxin Li** [2], **Duojie Weng** [2], **Wu Chen** [2] *

[1] School of Computer Science, China University of Geosciences, Wuhan 430074, China
[2] Department of Land Surveying and Geo-Informatics, The Hong Kong Polytechnic University, Hong Kong 999077, China
* Corresponding Author: wu.chen@polyu.edu.hk

**Abstract:** 3D reconstruction plays an increasingly important role in modern photogrammetric systems. Conventional satellite or aerial-based remote sensing (RS) platforms can provide the necessary data sources for the 3D reconstruction of large-scale landforms and cities. Even with low-altitude UAVs (Unmanned Aerial Vehicles), 3D reconstruction in complicated situations, such as urban canyons and indoor scenes, is challenging due to the frequent tracking failures between camera frames and high data collection costs. Recently, spherical images have been extensively exploited due to the capability of recording surrounding environments from one camera exposure. Classical 3D reconstruction pipelines, however, cannot be used for spherical images. Besides, there exist few software packages for 3D reconstruction of spherical images. Based on the imaging geometry of spherical cameras, this study investigates the algorithms for the relative orientation using spherical correspondences, absolute orientation using 3D correspondences between scene and spherical points, and the cost functions for BA (bundle adjustment) optimization. In addition, an incremental SfM (Structure from Motion) workflow has been proposed for spherical images using the above-mentioned algorithms. The proposed solution is finally verified by using three spherical datasets captured by both consumer-grade and professional spherical cameras. The results demonstrate that the proposed SfM workflow can achieve the successful 3D reconstruction of complex scenes and provide useful clues for the implementation in open-source software packages. The source code of the designed SfM workflow would be made publicly available[1].

## 1. Introduction

3D reconstruction plays an increasingly critical role in modern photogrammetric systems, which has been widely utilized for building modeling (Xiong et al., 2015), emergency response (Vetrivel et al., 2015), transmission corridor inspection (Jiang and Jiang, 2019; Jiang et al., 2017), etc. 3D reconstruction can be implemented by using data sources from varying sensors, such as optical cameras and laser scanners. Due to the low economic costs and the mature of image processing techniques, perspective cameras are the most popular sensors for 3D modeling, which have been equipped with different remote sensing (RS) platforms that range from high-altitude satellites (Yu et al., 2021) to low-altitude UAVs (Unmanned Aerial Vehicles) (Jiang et al., 2020; Jiang et al., 2022a).

---

[1] https://github.com/json87/SphereSfM

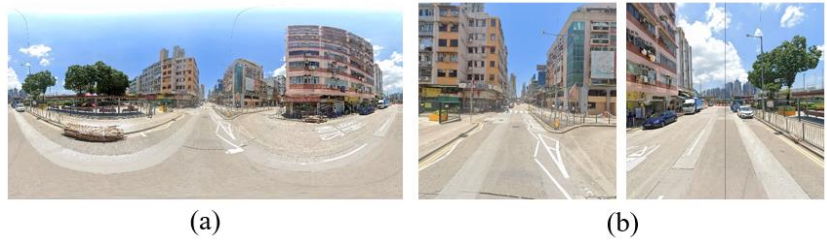

**Figure 1**. The illustration of the comparison between spherical and perspective images: (a) spherical image with full FOV; (b) perspective images with 45 degrees FOV.

Except for 3D reconstruction of large-scale terrains and urban buildings, recent years have witness the increasing demand for 3D modeling of fine-scale targets, e.g., building facades and urban streets. Although flexible data acquisition can be achieved through UAV-based nap-of-the-object or optimized views photogrammetry (Li et al., 2023), the observation ability of aerial RS platforms is still unsatisfactory in complex urban environments. To cope with this situation, mobile mapping systems (MMS) are further exploited in the urban environments. However, perspective cameras cannot adapt well to the characteristics (da Silveira et al., 2022) of the data acquisition in the complex environments mainly because of their limited FOV (Field of View). This can be explained from two aspects. On the one hand, the trajectory of ground vehicles would be seriously restricted by street structures, while the trajectory can be adjusted flexibly for aerial RS platforms. This would cause sudden viewpoint changes at street turning points and frequent track failure between camera frames (Ji et al., 2020). On the other hand, the image observation regions are expanded from the single direction in the aerial RS platforms to the full direction in the ground RS platforms, which requires more image recording at each camera exposure position and increases the acquisition time consumptions (Zhang et al., 2016). Thus, efficient imaging techniques are required for 3D reconstruction.

Recently, spherical cameras that have 360 and 180 degrees FOV in horizontal and vertical directions are featured as recording full surrounding environments from one camera exposure. **Figure 1** presents an illustration of the comparison between spherical and perspective images. It is clearly shown that spherical images can cover the whole scene from each camera record when compared with perspective images with very limited FOV. In addition, consumer-grade spherical cameras with low costs are becoming more and more popular, e.g., the Insta360 and Ricoh Theta (Gao et al., 2022). The capability and popularity of spherical cameras facilitate data acquisition in urban scenes and promote their usage for varying applications, including but not limited to underwater collision detection (Li et al., 2022), damaged building evaluation (Jhan et al., 2022), and urban 3D modeling (Fangi et al., 2018). Spherical images have become one of the most popular RS data sources for 3D reconstruction of urban scenes.

Compared with perspective images, there is a major difference for spherical images in the context of 3D reconstruction, i.e., the camera imaging model (Pagani and Stricker, 2011). In contrast to the 2D plane imaging of 3D points in the perspective projection, spherical camera imaging model projects 3D points onto the 3D sphere. For the image storage, the 3D sphere are then flattened onto the 2D image plane, which causes serious geometric distortions (Jiang et al., 2023). The difference of camera imaging model requires extra consideration for 3D modeling. In the literature, some research has been designed and reported to achieve 3D modeling from spherical images. To alleviate the distortion in feature matching, both image rectification and redesigned algorithms have been proposed (Chuang and Perng, 2018; Guan and Smith, 2017; Taira et al., 2015; Wang et al., 2018; Zhao et al., 2015). Similar to feature matching for oblique images, Wang et al. (2018) proposed converting spherical images into the cubic-map format that consists of six perspective images and casts feature matching from spherical images to the traditional perspective images. Considering few distortions exist in the sphere equator, Taira et al. (2015) rotated spherical images and detected local features from regions near the equator, which can be seen as a semi-global rectification method when compared with the cubic-map conversion. By using local rectification, Chuang and Perng (2018) proposed reprojecting the local image patches of keypoints onto the corresponding tangent planes and calculating feature descriptors from the rectified image patches. In contrast to the rectification-based methods, other researchers focus on redesigning the algorithms, such as the SPHORB (Zhao et al., 2015) and BRISKS (Guan and Smith, 2017).

To achieve the 3D reconstruction of spherical images, other attempts have also been made from earlier two-view or multi-view image orientation (Torii et al., 2005) to recent large-scale 3D modeling based on Structure from Motion (SfM) (Jhan et al., 2022; Zhang et al., 2020). In the work of Torii et al. (2005), both two and three-view geometry were presented, which establishes the basic geometry for 3D modeling of spherical images. As the pioneer work, (Fangi, 2007; Fangi and Nardinocchi, 2013; Pagani and Stricker, 2011) proposed the concept of spherical photogrammetry (SP) in the field of photogrammetry and remote sensing and established the workflow for 3D modeling of cultural heritage documentation. By using Google Street Views, (Micusik and Kosecka, 2009; Torii et al., 2009) implemented 3D modeling of urban streets and verified the usage of spherical images for urban street reconstruction. By investigating the use of spherical images in Structure from Motion, Pagani and Stricker (2011) designed error models for the relative and absolute orientation of spherical images. All these above-mentioned work has promoted the development of 3D reconstruction of spherical images.

However, there are only fewer open-source and commercial software packages that are designed for 3D reconstruction of spherical images when compared with perspective images. As far as we know, the satisfied software packages include MicMac, OpenMVG, Pix4dMapper, and Agisoft Metashape (Jiang et al., 2023). With the popularity of spherical cameras, such as consumer-grade Insta360, Ricoh Theta, and the development of image processing technology, the demand for 3D reconstruction of spherical images would increase dramatically. Therefore, this study aims to give an implementation of 3D reconstruction workflow for spherical images based on an incremental SfM (Structure from Motion) engine. The major contributions of this study include: (1) we present the basic camera imaging model for spherical cameras and give an insight analysis of key techniques for SfM-based 3D reconstruction of spherical images; (2) we implement a 3D reconstruction workflow based on an incremental SfM engine for spherical images, including the modules of feature matching, image orientation, cubic-map conversion, and dense matching; and (3) we verify the validation of the proposed SfM workflow by using spherical images recorded by both consumer-grade and professional cameras.

This paper is organized as follows. Section 2 presents the spherical image representation and camera model. Section 3 presents the key techniques for SfM-based 3D reconstruction. Section 4 proposes a 3D reconstruction pipeline for spherical images through incremental SfM-based image orientation and cubic-map-based dense matching, which is followed by the tests presented in Section 5. Finally, Section 6 concludes the work and future studies.

**2. Spherical camera model**

Camera model is the core to establish the projection relationship between 3D points in the object space and corresponding points in the image plane. It is the core module to implement 3D reconstruction by using spherical images. In this subsection, the commonly utilized image representation and camera model would be presented.

*2.1. Image representation*

Spherical images can record the full surrounding environments from one camera exposure. According to the purpose of different usage, there are three formats for image representation of spherical images, as illustrated in **Figure 2**. The first one is the direct spherical representation, in which 3D points in the object space are projected onto 3D points on the sphere, as presented in **Figure 2**(a). Spherical representation has the advantages for panoramic navigation that has been extensively adopted in well-known street-view navigation services, e.g., the Google and Baidu street-view maps. To facilitate image processing and hardware storage, equirectangular representation is the second format that is implemented through the equirectangular projection of 3D spherical images to 2D planar images, as shown in **Figure 2**(b). Compared to perspective images, equirectangular images can be processed directly by using existing algorithms, e.g., SIFT for feature extraction and matching (Pagani and Stricker, 2011). However, because of the projection from 3D sphere to 2D plane, serious geometric distortions are introduced to the regions that are far from the sphere equator in the equirectangular representation. To alleviate the distortions, cubic-map representation is the third image representation that converts each spherical image into six perspective images, as shown in **Figure 2**(c). Since the normal format, equirectangular representation (ERP) has been extensively adopted for spherical images and used in open-source and commercial software packages, including OpenMVG, Pix4dMapper, and Agisoft Metashape. Therefore, this study focuses on the equirectangular representation of spherical image in the following sections.

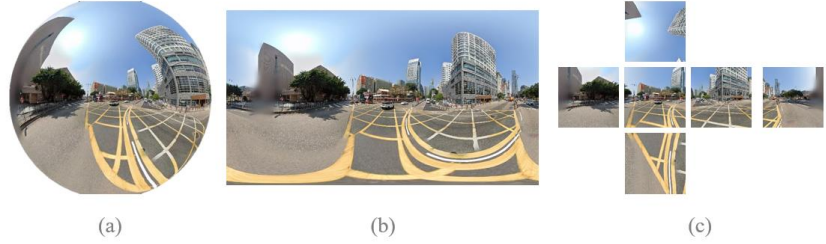

**Figure 2.** Three typical formats for spherical image representation: (a) spherical representation; (b) equirectangular representation; (c) cubic-map representation (Jiang et al., 2023).

*2.2. Camera model*

In the literature, there are three major camera models, i.e., the unified camera model, general camera model, and multi-camera model. In the unified camera model (Mei and Rives, 2007), environment light rays intersect into a single point, i.e., the projection center of the mirror, as shown in **Figure 3**(a). The unified camera model obeys the theoretical projection that can model the real-world imaging errors. On the contrary, the general camera model (Scaramuzza et al., 2006) uses the Taylor polynomial function to fit the theoretical projection, which can adapt to varying spherical cameras. For the multi-camera rig, the multi-camera model has been designed to establish the projection of multi-camera sensors, which can be implemented by an individual camera model or a unit sphere camera model (Ji et al., 2014). The individual camera model is rigorous in formulating the imaging system as shown in **Figure 3**(b); the unit sphere camera model simplifies camera projection by using a straightforward formula as shown in **Figure 3**(c), which has also been used in the open-source and commercial software packages, e.g., OpenMVG and Pix4dMapper.

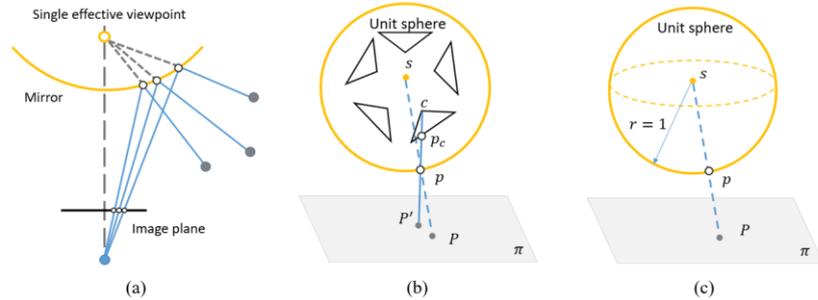

**Figure 3.** Three typical camera models: (a) unified camera model; (b) individual camera model; and (c) unit sphere camera model (Jiang et al., 2023).

Recently, the widely used spherical images are often collected by using a combination of multiple fisheye camera lenses. For example, the consumer-grade camera Insta360 and the professional-grade camera Ladybug use the combined panoramic cameras with 2 and 6 fisheye lenses, respectively. According to the comparative analysis of the camera models (Ji et al., 2014) and considering the versatility of the SfM workflow, this study adopts the unit sphere camera model to establish the intrinsic imaging model of spherical cameras. The intrinsic parameters $K$ of a spherical camera include the focal length $f$ and the principal point $(c_x, c_y)$. For an unit sphere camera model with the radius $r = 1$, the focal length of the spherical camera is $f = 1$; the principal point coordinates are $c_x = W/2$ and $c_y = H/2$, in which $W$ and $H$ indicate the image width and height, respectively.

## 3. 3D reconstruction workflow of spherical images

The workflow for 3D reconstruction of spherical images is designed as shown in Figure 4. The inputs are spherical images stored in equirectangular representation, and the outputs are dense point clouds, which can be further processed for textured models. In the workflow, there are three major components, i.e., image matching, image orientation, and dense matching. The key techniques involved in the SfM-based workflow are described as follows.

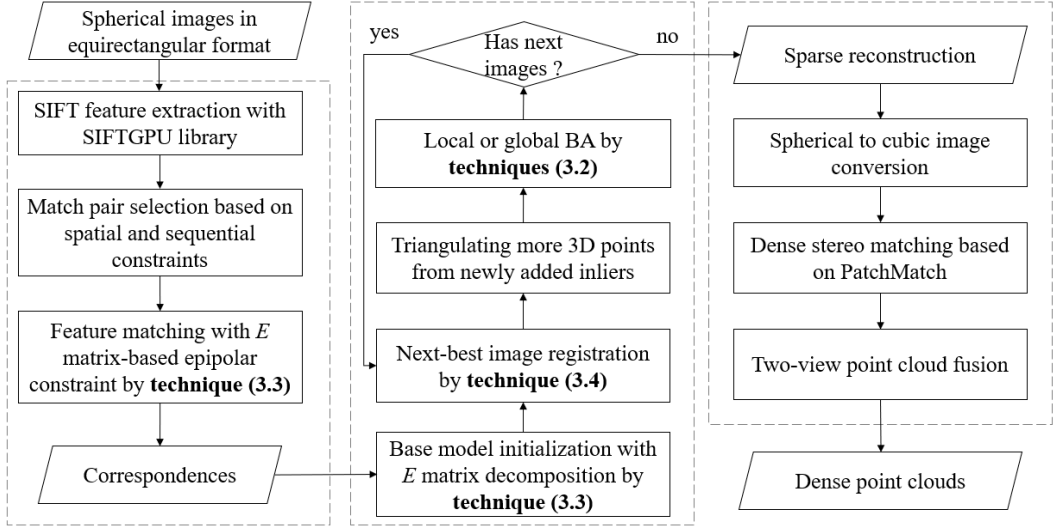

**Figure 4.** The 3D reconstruction workflow of spherical images.

*3.1. Camera imaging model*

Camera imaging model establishes the geometric transformation between 3D points in the object space and 2D points in the ERP image plane. The camera imaging model of spherical images is presented in **Figure 5**, in which **Figure 5**(a) presents the transformation between one 3D point $P$ in the object space and its corresponding 3D point $p$ on the sphere, and **Figure 5**(b) shows the transformation between the 3D point $p$ and its corresponding 2D point on the ERP image plane. For the spherical image in **Figure 5**(a), there are two coordinate systems, i.e., the spherical geographic coordinate system $O - r\theta\varphi$ and spherical Cartesian coordinate system $O - XYZ$. In the spherical geographic coordinate system, the coordinate of point $p$ is represented by using the longitude $\theta$ and latitude $\varphi$; in the spherical Cartesian coordinate system, the coordinate of point $p$ is presented by three coordinate terms $(x, y, z)$.

Suppose that the line $L$ shown by the dashed line in **Figure 5**(a) is the intersection of the Equatorial plane and the geodesic plane that passes point $p$. The longitude $\theta$ and latitude $\varphi$ are defined as the intersection angle between $L$ and $Z$ axis and the intersection angle between $L$ and $OP$, respectively; 3D point $P$ are projected onto 3D sphere point $p = (x, y, z)^T$. The transformation between the coordinate system $O - r\theta\varphi$ and $O - XYZ$ can be expressed using Equation (1), in which the sphere radius $r$ is set as one for the unit sphere camera model.

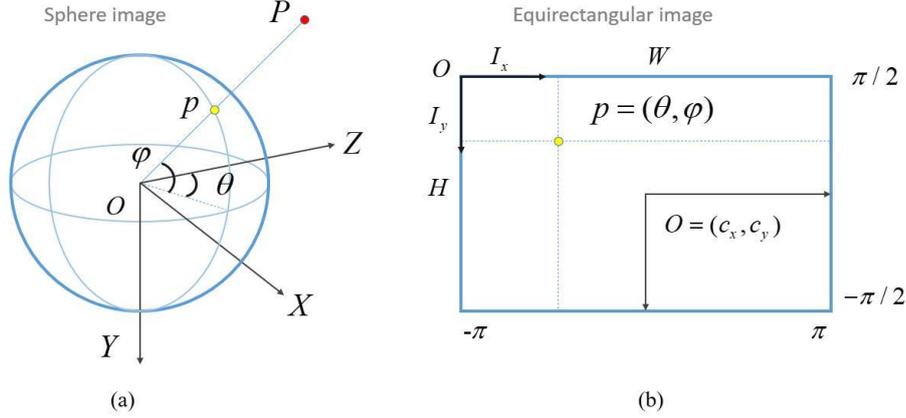

**Figure 5.** The principle of spherical camera imaging model and the coordinate transformation between the spherical image and equirectangular image: (a) camera imaging model; (b) image coordinate transformation between the spherical image and equirectangular image.

$$\begin{pmatrix} x \\ y \\ z \end{pmatrix} = \begin{pmatrix} \cos(\varphi)\sin(\theta) \\ -\sin(\varphi) \\ \cos(\varphi)\cos(\theta) \end{pmatrix} \quad (1)$$

According to the relationship between the 3D spherical geographic coordinate system and the 2D ERP image plane coordinate system as illustrated in **Figure 5**(b), their transformation can be expressed by using Equation (2)

$$\begin{pmatrix} \theta \\ \varphi \end{pmatrix} = \begin{pmatrix} \dfrac{I_x - c_x}{W} * 2\pi \\ \dfrac{c_y - I_y}{H} * \pi \end{pmatrix} \quad (2)$$

where $I_x$ and $I_y$ are the image coordinates in the ERP image plane; $c_x$ and $c_y$ are the coordinates of the origin $O$ of the ERP image plane; $W$ and $H$ are the image width and image height. Equations (1) and (2) establish the transformation between the 2D image plane coordinate system and 3D spherical coordinate system. Suppose that the pose of one spherical image is represented by the rotation matrix $R$ and the translation vector $T$ with respect to the world coordinate system $O - X_W Y_W Z_W$, one 3D point $P_W$ in $O - X_W Y_W Z_W$ can be transformed into the 3D spherical point $p$ in $O - XYZ$ by using Equation (3). Thus, Equations (1), (2), and (3) consist of the imaging model of spherical cameras.

$$\begin{cases} P = R * P_W + T \\ p = P / \|P\| \end{cases} \quad (3)$$

*3.2. Cost functions for bundle adjustment*

Cost functions are used to measure the residuals during refining the initial estimation of unknown parameters. For spherical image orientation, there are two kinds of cost functions. The first one is used to measure the transformation error $C_{trans}$ between two 3D spherical points $p_1 = (x_1, y_1, z_1)^T$ and $p_2 = (x_2, y_2, z_2)^T$ under the estimated transformation parameters, which is mainly utilized for the relative pose estimation based on spherical 3D correspondences. In this study, the cost function $C_{trans}$ is expressed according to Equation (4)

$$c_{trans} = \frac{(p_2^T E p_1)^2}{\|E p_1\|^2 + \|E p_2\|^2} \quad (4)$$

where $E$ indicates the estimated transformation between two spherical images; $\|\cdot\|$ represents the vector length. The second one is used to measure the reprojection error $C_{rprj}$ from 3D scene points $X_i$ to camera $C_j$, which would be used in the PnP (Perspective-n-Point) based absolute pose estimation as well as the local and global BA optimization. The cost function $C_{rprj}$ is represented by Equation (5)

$$c_{rprj} = P(C_j, X_i) - x_{ij} \quad (5)$$

where $P(C_j, X_i)$ represents the reprojection point $x'_{ij} = (I_x, I_y)$ from 3D point $X_i$ to camera $C_j$, which is calculated by using Equations (1)-(3); $x_{ij}$ is the image observation corresponding to 3D point $X_i$. These two cost functions would be used for parameter optimization.

*3.3. Relative orientation using spherical correspondences*

Relative orientation recovers the relative rotation $R$ and translation $T$ of two images by using the coplanar constraint of corresponding rays. Similar to the plane perspective imaging, spherical imaging still maintains the colinear equation, in which the projection center, image point and object point are collinear. Thus, the coplanar constraint is also appliable to spherical images. Suppose that the camera intrinsic parameters $K$ are known, the relative orientation parameters can be expressed as the essential matrix $E$. For two corresponding rays $p_1$ and $p_2$, they satisfy the coplanar constraint as shown in Equation (6)

$$p_2^T E p_1 = 0 \quad (6)$$

where $p_1$ and $p_2$ are the spherical coordinates of two corresponding image point $x_1$ and $x_2$, which are calculated according to the Equations (1) and (2). The geometrical meaning of the essential matrix $E = [T]_\times R$ is illustrated in Figure 6. When $p_1$ and $p_2$ are true corresponding points, the vectors $Rp_1$, $p_2$ and $T$ are coplanar. In other words, $p_2$ lies on the circular plane composed of the vector $Rp_1$ and $T$ (the normal vector of the circular plane is $\vec{n}$). In this study, the 8-point algorithms (Hartley and Zisserman, 2003) is adopted to solve the $E$ matrix, in which eight corresponding points form eight linear equations, and the linear system is solved through SVD (Singular Value Decomposition) (Umeyama, 1991).

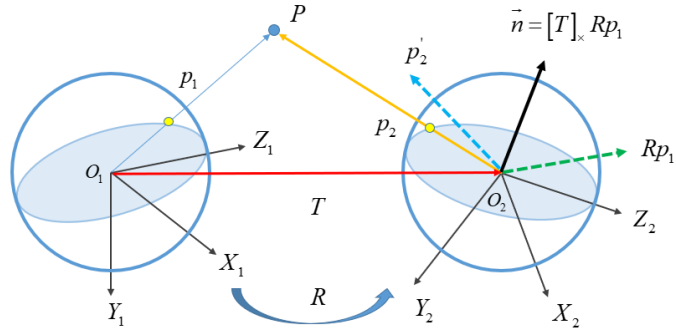

Figure 6. The principle of relative orientation for spherical images.

Due to outliers in initial corresponding matches, the hypothesis-verify framework based on RANSAC (Fischler and Bolles, 1981) has been utilized for robust estimation, which depends on the error metric $e$ and error threshold $e_p$. Different from the point-to-line perpendicular distance for perspective images, the corresponding ray $p_2$ of $p_1$ in the left image $O_1$ lies on the circular plane that is defined by the normal vector $\vec{n}$ and the projection center $O_2$. Thus, this study adopts the vector-to-plane geodesic angular error metric (Pagani and Stricker, 2011), which is formulated according to Equation (7)

$$e = abs\left(\sin^{-1}\left(p_2^T E p_1\right)\right) \tag{7}$$

where $abs(\cdot)$ indicates the absolute value. On the other hand, the unit of the error threshold $e_p$ in image pixels should be converted to spherical angles that are used in the metric $e$. In this study, the conversion is achieved based on Equation (8)

$$e_a = \frac{2\pi}{\max(W, H)} e_p \tag{8}$$

where $2\pi/\max(W, H)$ indicates the scale factor of these two metrics; $e_a$ is the error threshold in spherical angles. Thus, for an estimated matrix $E$, the corresponding points $p_1$ and $p_2$ are labeled as one inlier if their angular error $e < e_a$.

After the robust estimation based on RANSAC, four possible solutions can be obtained by the decomposition of the essential matrix $E$, and the cheirality check is then conducted to select the right solution. Instead of restricting the depth of resumed 3D scene points to be positive for perspective images, a consistent direction between 3D scene points $P$ and 3D spherical points $p$ is checked in the cheirality check. This constraint is formulated by $p^T(RP + T) > 0$. The solution with the largest number of consistent 3D scene points is selected as the initial $R$ and $T$, which are then refined in BA optimization with the cost function in Equation (4).

*3.4. Absolute orientation using 3D correspondences*

The relative pose and 3D scene points can be recovered based on the relative orientation and the triangulation of spherical correspondences. In perspective imaging, the absolute pose of newly added images can be directly computed based on the 2D-3D correspondences using the PnP (Perspective-n-Point) algorithm. Since the colinear relationship among the projection center, image points and scene points is still satisfied, the PnP can also be utilized for absolute orientation, which is implemented by using the collinear constraint of 3D spherical point $p$ and 3D scene point $P$, as presented in Equation (9), in which $R$ and $T$ are the rotation matrix and translation vector of the spherical image. In this study, an initial solution is calculated by using three correspondences based on the P3P algorithm (Gao et al., 2003).

$$[p]_\times (RP + T) = 0 \tag{9}$$

To cope with outliers, the RANSAC algorithm is adopted for robust estimation. In contrast to the pixel distance residuals for perspective images, the angular error metric $e$ is used to calculate the reprojection error as shown by Equation (10)

$$e = abs\left(\cos^{-1}\left(p^T(RP + T)\right)\right) \tag{10}$$

where $RP + T$ indicates the projection point of the 3D scene point $P$ in the camera system; $e$ is the angle between $p$ and $RP + T$. Similar to relative pose estimation, the error threshold $e_a$ is also calculated by using Equation (8), and the correspondences are labeled as inliers if their error $e < e_a$. After obtaining the initial solution, BA optimization is then executed by using the cost function represented in Equation (5), which minimizes the reprojection error.

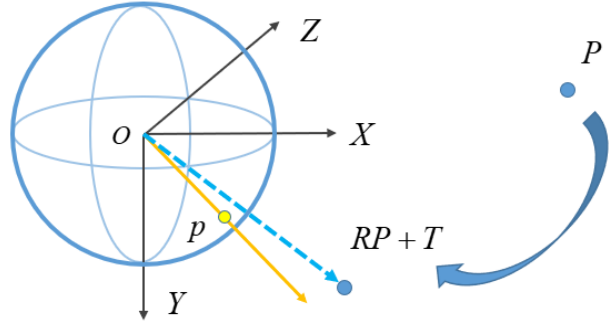

Figure 7. The principle of absolute orientation error for spherical images.

**4. Implementation of the 3D reconstruction workflow**

In the SfM-based workflow, the used key techniques are described in Section 3, and three components, i.e., image matching, image orientation, and dense matching are designed and implemented as illustrated in Figure 4.

**Image matching** is used to establish reliable matching points of two-view images, which is the basis of SfM-based 3D reconstruction. The pipeline of image matching includes three major steps, including feature extraction, feature matching and outlier removal.

(1) Feature extraction. Although geometric distortions are introduced in the ERP projection, this study uses the classic SIFT algorithm (Lowe, 2004) for feature extraction considering two main reasons. On the one hand, a majority of spherical images are recorded by fixing the camera roll (around Z axis) and pitch (around X axis) angles, such as the ground MMS system, which ensures that the consistency of geometric structures near the equator of the ERP images; on the other hand, SIFT is invariant to rotation and scale as well as has a high tolerance to the changes of imaging viewpoints and illuminations. In this implementation, the high-performance open-source library SIFTGPU (Wu, 2007) with default parameter settings has been used for feature extraction.

(2) Feature matching. Before feature matching, effective matching pair selection can reduce the high computational costs of exhaustive matching and avoid introducing false matches. Considering that most spherical cameras have built-in GNSS (Global Navigation Satellite System) sensors that can record the recording locations of images and at the same time, spherical images are usually acquired sequentially, this study uses the spatial constraint and sequential constraint to select matching pairs and guide feature matching. For feature matching, the SIFTGPU library is used by setting the ratio test and the maximum distance threshold with the values of 0.8 and 0.7, respectively.

(3) Outlier removal. SIFT uses the local image patches around keypoints to calculate feature descriptors, which would inevitably introduce outliers in the initial matches. Based on RANSAC-based essential matrix estimation as presented in Section 3.3, this study uses the angular error metric $e$ between a 3D spherical point and its corresponding circular plane to eliminate outliers and obtain refined matches. In the $E$ matrix estimation based on the RANSAC, the error threshold $e_p$ is set as 4 pixels.

**Image orientation** is implemented based on an incremental SfM engine by using reliable matches, which iteratively solves and optimizes image poses and 3D points. The main steps of the workflow include seed image reconstruction, absolute orientation of the next-best image, and local or global BA optimization.

(1) Seed image reconstruction. The initial seed images construct the basic model for the entire incremental SfM reconstruction. Seed image selection should consider both the number of matches and the intersection angle of images. The main steps include: (1) sort images in the descending order according to the number of matches; (2) select the first image in the ordered sequence as the first image $I_{first}$ in the seed images; (3) sort associated images of $I_{first}$ in the descending order according to the number of matches; (4) select the first image from the ordered associated image as the second image $I_{second}$ in the seed images; (5) conduct the relative orientation of these two seed images using spherical correspondences according to Section 3.3; (6) Iteratively execute steps (2)-(5) until two seed images are found, which satisfy the threshold of the number of matches $N_{inlier}$ and the intersection angle $A_{tri}$. In this study, $N_{inlier} > 100$, $A_{tri} > 16°$. After the basic model construction by using seed images $I_{first}$ and $I_{second}$, the global BA optimization is performed according to the cost function represented by Equation (5).

(2) Absolute orientation of the next-best image. The next-best image represents the most robust candidate image that can be registered to the reconstructed model. The selection of the next-best image is based on the number of observed 3D points and the image plane distribution of corresponding 2D feature points: (1) for all unoriented images, build the mapping relationships between feature points and reconstructed 3D points; (2) filter the images whose match number $N_{obs} < 30$ and calculate the important value $Score$ based on the image plane distribution of corresponding 2D feature points as described in the work of Schonberger and Frahm (2016); (3) sort remaining images in the descending order of $Score$, and the candidate image with the highest score is the next-best image; (4) using the RANSAC-based absolute pose estimation as presented in Section 3.4, the rotation $R$ and position $T$ of the next-best image are calculated, which are then refined in BA optimization according to cost function in Equation (5); (5) the inliers of the next-best image are triangulated to resume more 3D scene points.

(3) Local or global BA optimization. After the successful adding of the new next-best image, local or global BA optimization is executed according to two conditions: (1) the number of newly added images $N_{i\_add} > 3$, local BA optimization is executed to refine only the poses of newly added images and their associated 3D points; (2) the number of newly added images $N_{i\_add}$ or the number of newly added 3D points $N_{p\_add}$ is greater than a given threshold, global BA optimization is executed to optimize all images and 3D points. In this paper, the thresholds of $N_{i\_add}$ and $N_{p\_add}$ are set to 10% of the number of reconstructed images and 3D points. According to the cost function in Equation (5), the goal is to minimize the reprojection error of 3D points

$$\min_{C_j, X_i} \sum_{i=1}^{n} \sum_{j=1}^{m} \rho_{ij} \left\| P(C_j, X_i) - x_{ij} \right\|^2 \qquad (11)$$

where $\|\cdot\|$ represents the vector L2 norm; $\rho_{ij}$ is the visibility indicator of a 3D point $X_i$ in the image $C_j$. When $X_i$ is visible in image $C_j$, $\rho_{ij} = 1$; otherwise, $\rho_{ij} = 0$.

**Dense matching** is used to generate point clouds from the result of SfM reconstruction. To make full use of existing dense matching algorithm, this study first converts the SfM results of spherical images into the format in the cubic-map image representation.

(1) Cubic-map image generation. Assuming that the internal orientation matrix of the cubic-map image $I$ is represented by $K_P$, and its rotation matrix to the coordinate system of the spherical image is $R_{PS}$, the cubic-map image is generated with the following steps: (1) for an image point $x \in I$, calculate its image coordinate and then perform homogeneous normalization by using $u = \prod K_P^{-1} x$, in which $\prod$ represents homogeneous normalization; (2) convert the homogeneous normalized coordinate $u$ to the spherical coordinate system and obtain the corresponding spherical Cartesian coordinate $u' = R_{PS}^T u$; (3) according to Equations (1) and (2), calculate the pixel coordinate $x'$ in the ERP image for $u'$, and linearly interpolate the gray value for the cubic-map image point $x$. Through the steps (1)-(3), the cubic-map image is generated.

(2) Pose update. Considering that the projection center of the cubic-map image coincides with that of the spherical image, the rotation matrix and translation vector should be updated by using the relative rotation matrix $R_{PS}$, the pose of the generated cubic-map image can be obtained by using Equation (12).

$$\begin{cases} R_P = R_{PS}R \\ T_P = -R_P\left(-R^T T\right) \end{cases} \quad (12)$$

(3) Dense matching. After the above pose transformation, this study uses a classic multi-view stereo (MVS) matching algorithm to generate dense point clouds. In the experiments, the PatchMatch (Schönberger et al., 2016) dense matching algorithm has been used.

## 5. Experiments and results

In the experiments, three spherical datasets are used for the performance evaluation of the proposed 3D reconstruction pipeline and comparison with other open-source and commercial software packages. First, we analyze the performance of the 3D reconstruction pipeline in terms of SIFT-based feature matching, SfM-based image orientation, and cubic-map converted dense matching. Second, three software packages that support spherical images are compared with the proposed 3D reconstruction, including the open-source package OpenMVG (Moulon et al., 2016) and commercial packages Pix4Dmapper (Pix4dMapper, 2022) and AgiSoft Metashape (Metashape, 2022). In this test, the proposed pipeline and OpenMVG are implemented by using the C++ programming language, and all experiments are conducted in a Windows desktop PC configured with 32 GB memory, an Intel Core i7-8700K 3.7GHz CPU (Central Processing Unit), and an NVIDIA GeForce GTX 1050Ti GPU (Graph Processing Unit).

*5.1. Test sites and datasets*

In this study, three spherical datasets are used to evaluate the performance of the proposed 3D reconstruction pipeline for spherical images. The detailed information for data acquisition and spherical images is presented in Table 1. The description of each dataset is listed as follows:

**Table 1.** Detailed information on the three spherical datasets.

| Item Name | Dataset 1 | Dataset 2 | Dataset 3 |
|---|---|---|---|
| Sensor type | Sphere | Sphere | Sphere |
| Camera mode | Garmin VIRB 360 | Garmin VIRB 360 | Ladybug3 |
| Number of cameras | 2 | 2 | 6 |
| Storage format | Equirectangular | Equirectangular | Equirectangular |
| Acquisition platform | Ground fixed | Hand held | Moving car |
| Number of images | 37 | 279 | 1937 |
| Image size (pixel) | 5640 × 2820 | 5640 × 2820 | 5400 × 2700 |

- The first dataset is collected from a campus site, which contains a parterre surrounded by teaching buildings, as illustrated in **Figure 8**(a). For image acquisition, a Garmin VIRB 360 camera with two fisheye sensors is utilized, which stores recorded spherical images in the equirectangular representation format. The data acquisition campaign is conducted around the central parterre, and there are a total number of 37 images collected from this test site, whose resolution is 5640 by 2820 pixels.
- The second dataset locates within a building dooryard that covers from its rooftop to the inner aisles, as shown in **Figure 8**(b). There are some parterres on the rooftop, and the inner aisles connect different layers. For data acquisition, the same Garmin VIRB 360 camera as used in dataset 1 has been adopted. To accelerate data acquisition in the complex buildings, the camera has been handheld instead of ground fixed in dataset 1, and there are 279 spherical images collected from these test sites, which cover whole inner aisles.
- The third dataset is collected from an MMS system that is mounted on a moving car. The test site goes straight along an urban road that is approximately 7 kilometers. Low urban buildings are located near the two sides of the urban road, as shown in **Figure 8**(c). In this test site, a Point Grey Ladybug3 camera that consists of six fisheye cameras is adopted for data acquisition, which stores images in the equirectangular format. By setting the interval value of 3 meters for camera exposure, there are a total number of 1937 spherical images collected from this site.

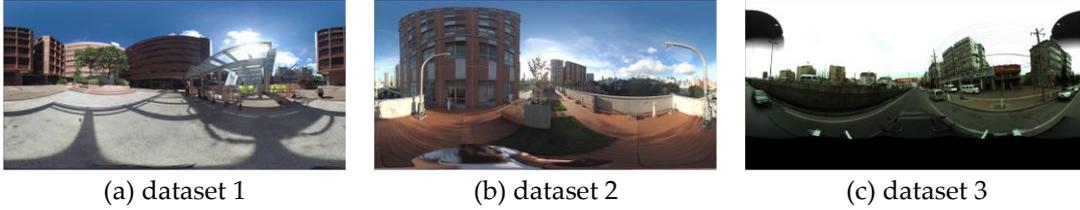

(a) dataset 1  (b) dataset 2  (c) dataset 3

**Figure 8**. The sample images from the three spherical datasets.

*5.2. Results of feature detection and matching*

Feature detection and matching are implemented by using the traditional SIFT algorithm. In this test, the GPU-based open-source library SIFTGPU has been used with default parameter configurations. For feature detection, the maximum number of detected features for each image is set as 8192, in which features with a larger scale would be retained. **Figure 9** illustrates feature detection results for one image in datasets 1 and 2, as presented in **Figure 9**(a) and **Figure 9**(b), respectively. The detected features are rendered as yellow circles. We can see that the detected features can cover the whole image, especially for building facades that could provide stable features for images with varying viewpoints. In addition, for the low-texture sky regions, there are only a small fraction of detected features that are near the boundaries of clouds, which helps to reduce the introduction of false matches in feature matching.

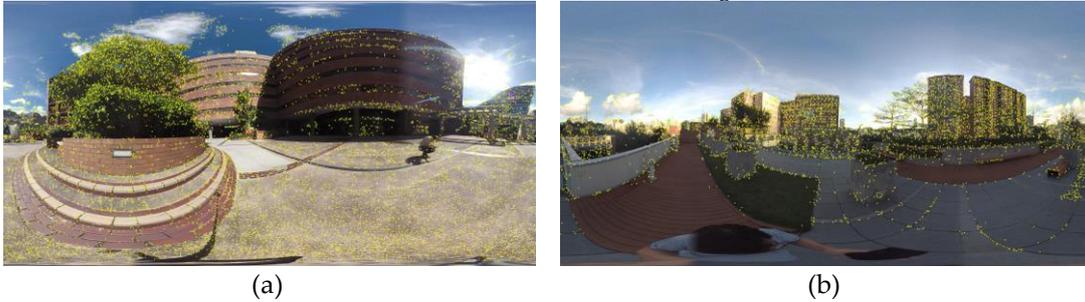

(a)  (b)

**Figure 9**. An illustration of detected features: (a) for one image in dataset 1; (b) for one image in dataset 2. The detected features are rendered as yellow circles.

Before feature matching, match pair selection is conducted to select image pairs that are spatially overlapped due to two reasons. On the one hand, it can filter unnecessary image pairs and decrease the time costs in feature matching; on the other hand, it can avoid the introduction of false matches that arise from non-overlapped image pairs. In practice, the strategy for match selection depends on the characteristics of data acquisition. For dataset 1, the exhaustive match pair selection strategy is used since images are recorded around a parterre. For datasets 2 and 3, the spatial and sequential constrained strategies are used as images are recorded with non-regular or corridor-like strategies, in which the maximum distance is set as 20 m in the spatial constraint, and the overlap image number is set as 10 in the sequential constraint. **Figure 10** shows the match pair selection results, in which match pairs are presented using TCN graph.

It is shown that enough connections can be established between images, and there are 157, 4941, and 14836 match pairs selected for the three datasets.

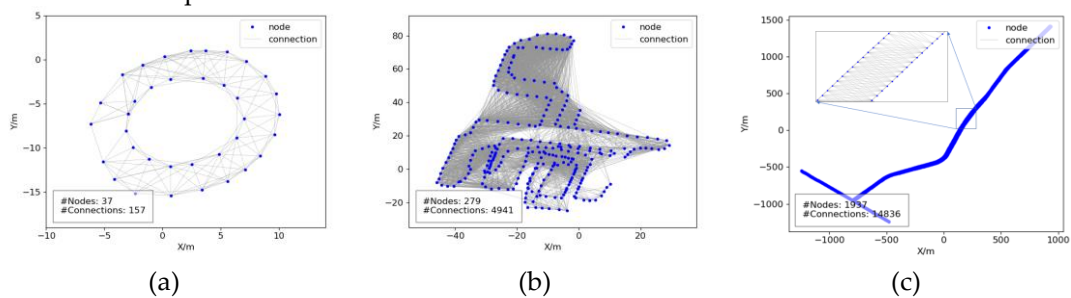

(a)　　　　　　　　　　　(b)　　　　　　　　　　　(c)

**Figure 10**. The match pair selection results presented by using image topological connection networks (TCN): (a) TCN for dataset 1; (b) TCN for dataset 2; (c) TCN for dataset 3. The number of selected match pairs is listed in the bottom left region.

Feature matching is then executed and guided by the selected match pairs, in which the distance ratio between the first and second closest descriptors and maximum distance between any two descriptors are set as 0.8 and 0.7, respectively. For outlier removal, cross-checking is used to reject false matches in the L2-norm-based initial feature matching, and essential matrix-based epipolar geometry is further used to remove remaining false matches, which is estimated through RANSAC with an error threshold configured as 4 pixels.

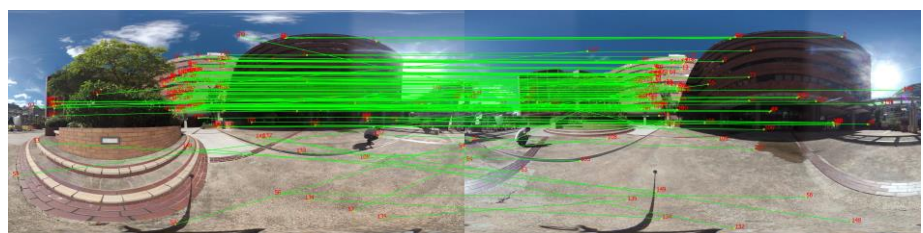

(a) dataset 1 – initial match (149)

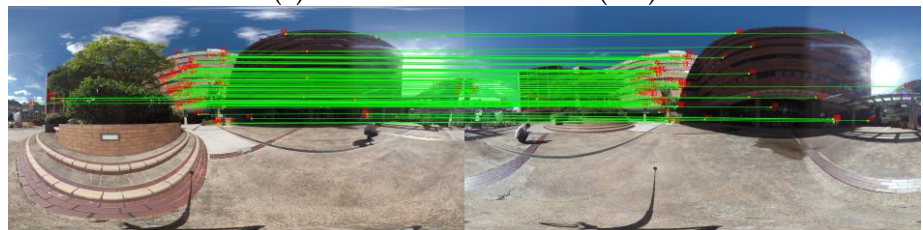

(b) dataset 1 – refined match (116)

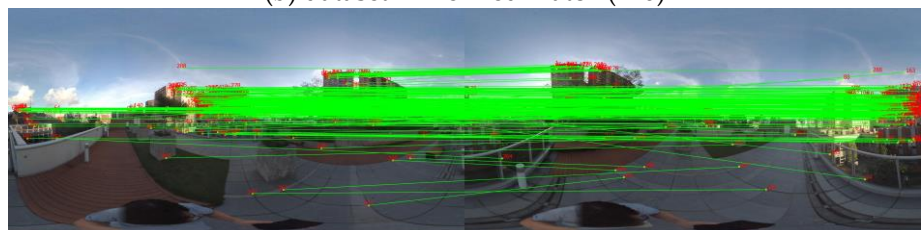

(c) dataset 2 – initial match (291)

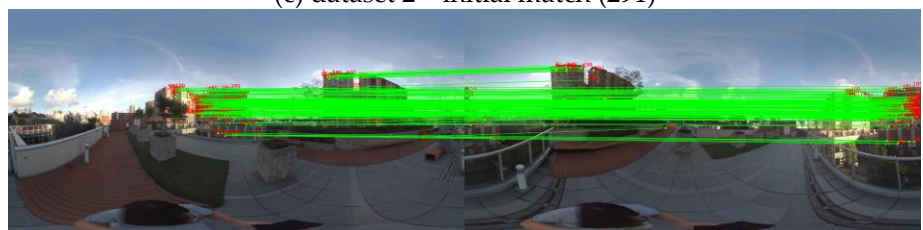

(d) dataset 2 – refined match (202)

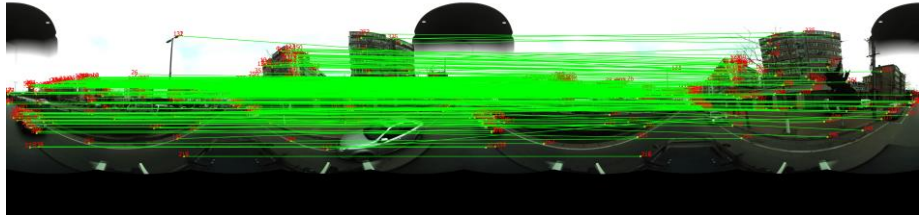

(e) dataset 3 – initial match (373)

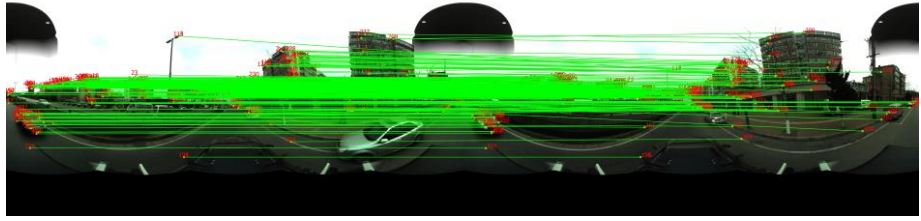

(f) dataset 3 – refined match (324)

**Figure 11**. Feature matching examples for three image pairs from the three datasets: (a) and (b) are the initial match and refined match for the image pair in dataset 1; (c) and (d) for the image pair in dataset 2; (e) and (f) for the image pair in dataset 3.

For visual analysis, **Figure 11** illustrates the examples of feature matching for three image pairs in the three datasets. Corresponding feature points are linked by using green lines; lines representing true matches are almost parallel; on the contrary, lines of false matches intersect others. It is shown that enough matches can be obtained from the used SIFT feature descriptors, and a majority of feature matches locate on building facades. For the three image pairs, there are a total number of 149, 291, and 373 initial matches. Due to repetitive patterns, false matches exist obviously in initial matches, especially for datasets 1 and 2. By using the essential matrix-based outlier removal strategy, these false matches can be filtered obviously. Finally, 116, 202, and 324 matches are retained for the three image pairs.

**Figure 12** illustrates the weight matrix of feature matching for the three datasets. The value in each cell indicates the number ratio between the feature matches of the corresponding match pair and the largest feature matches of all match pairs, which reflects the connection strength. We can see that all images have a strong connection with their neighboring images since larger values locate on the diagonal cells. In addition, extra connections have been established with other spatially closed images, which can be observed from the non-diagonal cells. In a word, all images are stringed into the image connection network.

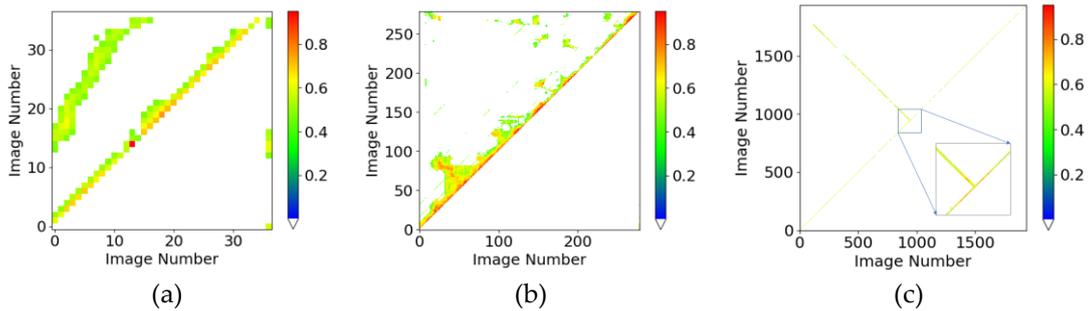

**Figure 12**. The weight matrix of feature matching for the three datasets: (a) weight matrix for dataset 1; (b) weight matrix for dataset 2; (c) weight matrix for dataset 3. The sub-figure in (c) indicates the zoomed content in the rectangular region.

### 5.3. Results of image orientation and dense matching

Image orientation is achieved based on an incremental SfM pipeline that aims to resume camera poses and sparse 3D points by using established two-view feature matches. In this test, three metrics are used to evaluate the performance of SfM-based image orientation, including efficiency, completeness, and precision. The metric efficiency indicates the time costs consumed in image orientation; the metric completeness indicates the number of resumed camera poses and 3D points; the metric precision represents the orientation accuracy, which is quantified by using the reprojection error in BA optimization.

**Table 2.** The statistical results of image orientation in terms of efficiency, completeness, and precision for the three datasets.

| Dataset | Efficiency (min) | Completeness | | Precision (pixel) |
| --- | --- | --- | --- | --- |
| | | 3D points | Images | |
| Dataset 1 | 0.25 | 3,044 | 37/37 | 0.786 |
| Dataset 2 | 12.16 | 38,995 | 279/279 | 0.798 |
| Dataset 3 | 140.50 | 290,262 | 1,937/1,937 | 0.581 |

Table 2 presents the statistical result of image orientation for the three datasets. We can see that all images can be successfully registered based on the proposed SfM solution, and there are 3044, 38995, and 290262 points generated for the three datasets, respectively. For the metric precision, the reprojection errors of the three datasets are almost consistent, which are 0.786, 0.798, and 0.581 pixels, respectively. Due to the usage of an incremental SfM solution, the time consumption increases dramatically with the increasing image numbers. For the three datasets, 0.25 min, 12.16 min, and 140.50 min are consumed in image orientation.

For visual analysis, **Figure 13** presents the sparse point clouds generated from datasets 2 and 3, in which images are rendered as blue rectangles. For dataset 2 as shown in **Figure 13**(a), spherical images are captured from four floors that are connected by inner stairs as illustrated by the top left spherical image. Noticeably, this test site includes building roofs, inner corridors, and underground tunnels. Due to the full FOV of special images, a reliable image connection can be established in this complex test site, which ensures the success of image orientation. For dataset 3 as shown in **Figure 13**(b), the length of the entire trajectory is about 8 kilometers, and all images are connected in image orientation. The successful image orientation of these two sites demonstrates the validation of the proposed solution for spherical images.

Instead of revising existing or redesigning new algorithms, the proposed solution converts spherical images into perspective images according to the cubic-map representation. Existing dense matching algorithms can be directly applied to the generated perspective images. In the proposed 3D reconstruction pipeline of spherical images, dense point clouds are generated based on the PatchMatch-based stereo matching algorithm.

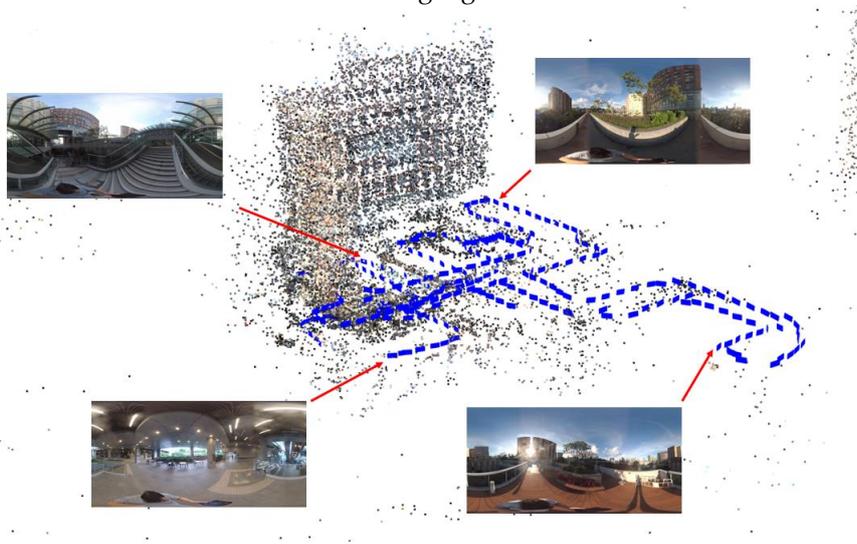

(a)

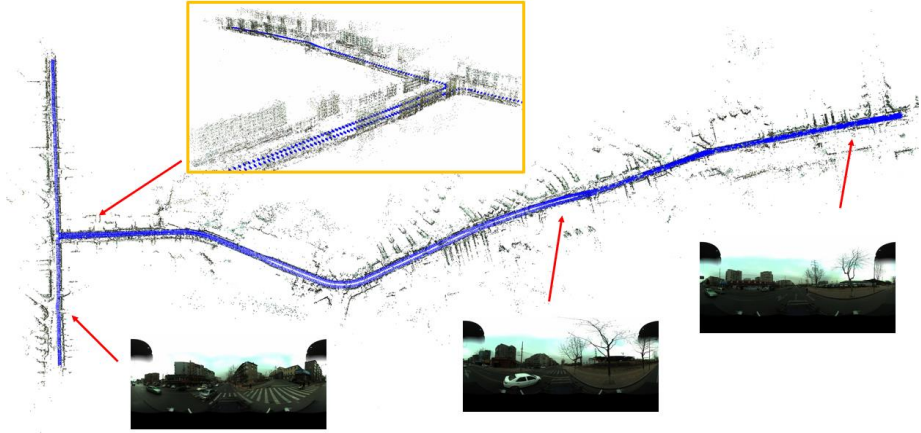

(b)

**Figure 13**. The illustration of SfM-based image orientation results: (a) result for dataset 2; (b) result for dataset 3. The blue rectangles represent image planes, and sparse 3D points are rendered by image colors.

**Figure 14** presents the dense point clouds generated from dataset 2. About 4.5 million 3D points are reconstructed from this test site. We can see that the overall structure of this test site has been successfully reconstructed, including building facades, underground corridors, and inner facilities, which are verified by the subfigures in **Figure 14**(a), **Figure 14**(b), and **Figure 14**(c). Especially for inner facilities, we can observe the overall layout and detailed structure of stairs and handrails, as presented in **Figure 14**(c). In addition, there is only one trajectory during data acquisition in underground corridors. The detailed structures of these regions verify the advantages of spherical images on 3D reconstruction in the indoor environment.

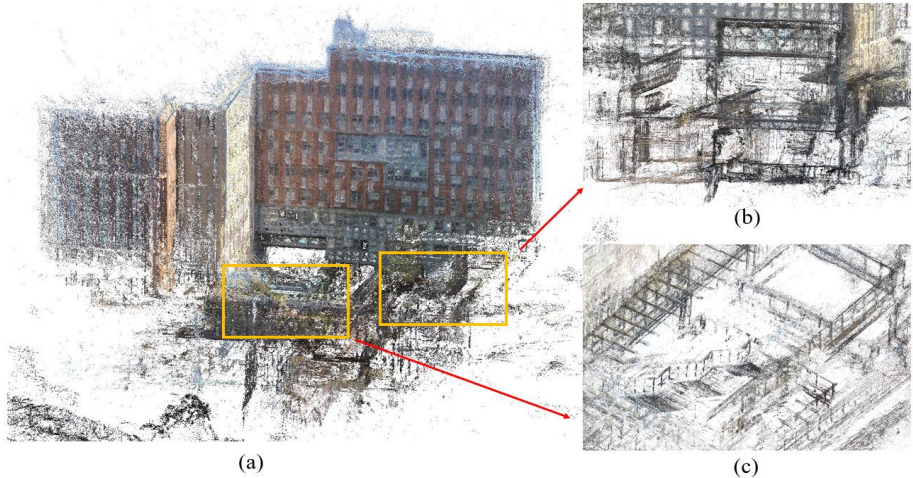

**Figure 14**. Dense matching point clouds of dataset 2: (a) the overall illustration; (b) details of the underground region; (c) details of the inner regions that include elevators and stairs.

*5.4. Comparison with other software packages*

Three software packages that support spherical images are evaluated and compared with the proposed 3D reconstruction pipeline. The selected software packages consist of one open-source software OpenMVG, and two commercial software Metashape and Pix4Dmapper. In this test, the performance of image orientation would be evaluated for two main reasons. On the one hand, it determines the success of 3D reconstruction; on the other hand, some software packages do not provide dense matching modules, e.g., OpenMVG. Table 3 presents detailed information for these three software packages.

**Table 3.** Detailed information for the compared software packages. The term POS indicates Positioning and Orientation System; sequential, multi-scale, and VOC-TREE indicate match pair selection based on the sequential constraint, multi-scale preemptive matching, and vocabulary tree based image retrieval, respectively; NNS represents Nearest Neighbor Searching, and ISfM indicates incremental Structure from Motion.

| Name | Image match pair | Feature matching | Image orientation | Version |
|---|---|---|---|---|
| OpenMVG | GNSS & sequential | CPU NNS | ISfM | V2.0 |
| Metashape | POS & multi-scale | GPU & CPU NNS | Cluster-based ISfM | 1.7 |
| Pix4Dmapper | GNSS & VOC-TREE | GPU & CPU NNS | ISfM | 4.4.12 |

Table 4 presents the statistical results of image orientation for the three datasets. It is shown that OpenMVG fails to reconstruct all three datasets. For datasets 1 and 2, the initialization of two seed images failed; for dataset 3, the subsequent SfM reconstruction failed. Compared with OpenMVG, the robustness of the other two commercial software packages Metashape and Pix4Dmapper is higher. Pix4Dmapper can register a majority of images for datasets 1 and 2 with the highest precision although the failure in dataset 3, whose precision is 0.359 pixels and 0.341 pixels for datasets 1 and 2, respectively. Metashape can register all images in the three datasets with the highest efficiency, which are 0.16 min, 1.20 min, and 8.90 min, respectively. The main reason is the usage of cluster-based parallel ISfM. Considering the proposed solution, we can see that it can register all images with higher precision when compared with Metashape; its efficiency, however, is lower, which are 0.25 min, 12.16 min, and 140.50 min, respectively. The main reason is the drawback of the used incremental SfM solution, which can be solved by using a divide-and-conquer strategy, such as our previous work (Jiang et al., 2022b). Thus, the comparison results reveal that 3D reconstruction of spherical images is an unsolved issue in both open-source and commercial software packages, and more attention should be paid to promoting the development of well-designed algorithms and toolkits.

**Table 4.** The statistical results of image orientation for the three datasets in terms of efficiency, completeness, and precision. The values in the bracket indicate the number of registered images in the image orientation result.

| Metric | Method | Dataset 1 | Dataset 2 | Dataset 3 |
|---|---|---|---|---|
| Efficiency (min) | OpenMVG | — | — | — |
| | Metashape | 0.16 | 1.20 | 8.90 |
| | Pix4Dmapper | 0.88 | 9.22 | — |
| | Ours | 0.25 | 12.16 | 140.50 |
| Precision (pixel) | OpenMVG | — | — | — |
| | Metashape | 5.260 | 4.450 | 0.958 |
| | Pix4Dmapper | 0.359 | 0.341 | — |
| | Ours | 0.786 | 0.798 | 0.581 |
| Completeness | OpenMVG | — | — | — |
| | Metashape | 5,540 (37) | 43,120 (279) | 295,390 (1,937) |
| | Pix4Dmapper | 8,485 (36) | 64,135 (238) | — |
| | Ours | 3,044 (37) | 38,995 (279) | 290,262 (1,937) |

## 6. Conclusions

Spherical images can record all surrounding environments by using one camera exposure. In contrast to perspective images with limited FOV, spherical images can cover the whole scene and have been increasingly used for 3D modeling in street-view and indoor environments. In this paper, we give a review of 3D reconstruction of spherical images according to the classical processing pipeline, i.e., feature detection and matching, image orientation to resume camera poses, and dense matching to generate point clouds. For feature detection and matching, the serious geometric distortions caused by equirectangular projection pose difficulties in feature description. Both traditional and newly designed algorithms to cope with distortions have been reviewed, as well as the deep learning-based methods. For image orientation, we reviewed the SfM-based offline methods and SLAM-based online methods and conducted a statistic of open-source and commercial software packages. For dense matching, we reviewed depth prediction from single images, traditional methods for cubic-map images, and recent deep learning CNN for two-view and multi-view stereo matching.

According to the differences between perspective and spherical images, we also proposed a 3D reconstruction pipeline through incremental SfM-based image orientation and cubic-map conversion-based dense matching. By using real spherical images recorded by both consumer-grade and professional spherical cameras, the validation of the proposed pipeline has been evaluated and compared with open-source and commercial software packages. The test results demonstrate that spherical images can be promising data sources for 3D reconstruction in complex environments, e.g., urban canyons and indoor corridors. In addition, the comparison with other software packages verified that the proposed pipeline is robust and effective for 3D reconstruction of spherical images, which provides clues to guide further research.

**Acknowledgment**

This research was funded by the National Natural Science Foundation of China under Grant [number 42001413] and the Hong Kong Scholars Program under Grant [number 2021-114].